\def\year{2020}\relax
\documentclass[letterpaper]{article} 
\usepackage{aaai20}  
\usepackage{times}  
\usepackage{helvet} 
\usepackage{courier}  
\usepackage[hyphens]{url}  
\usepackage{graphicx} 
\usepackage{graphicx}  
\usepackage{subcaption}
\usepackage{amsmath}

\DeclareRobustCommand\onedot{\futurelet\@let@token\@onedot}

\def\onedot{\ifx\@let@token.\else.\null\fi\xspace}

\def\etal{\emph{et al.}}

\title{AutoRemover: Automatic Object Removal for Autonomous Driving Videos}
\author{
Rong Zhang,\textsuperscript{\rm 1}\thanks{The authors from Zhejiang University are affiliated with the State Key Lab of CAD\&CG.}
Wei Li,\textsuperscript{\rm 2,4}$^\dagger$
Peng Wang,\textsuperscript{\rm 2}
Chenye Guan,\textsuperscript{\rm 2}
Jin Fang,\textsuperscript{\rm 2}  \\ \Large
\textbf{Yuhang Song,\textsuperscript{\rm 3}
Jinhui Yu,\textsuperscript{\rm 1}
Baoquan Chen,\textsuperscript{\rm 4}
Weiwei Xu,\textsuperscript{\rm 1}\thanks{Corresponding authors.}
Ruigang Yang,\textsuperscript{\rm 2}} \\
\textsuperscript{\rm 1}Zhejiang University, 
\textsuperscript{\rm 2}Baidu Research, Baidu Inc. \\
\textsuperscript{\rm 3}University of Southern California, 
\textsuperscript{\rm 4}Peking University \\
\small cadzhangrong@zju.edu.cn, liweimcc@gmail.com, jerryking234@gmail.com, \{guanchenye, fangjin\}@baidu.com, \\ 
\small yuhangso@usc.edu, jhyu@cad.zju.edu.cn, baoquan@pku.edu.cn, xww@cad.zju.edu.cn, ryang2@outlook.com
}

\begin{document}
\maketitle
\begin{abstract}
Motivated by the need for photo-realistic simulation in autonomous driving, in this paper we present a video inpainting algorithm \emph{AutoRemover}, designed specifically for generating street-view videos without any moving objects. In our setup we have two challenges: the first is the shadow, shadows are usually unlabeled but tightly coupled with the moving objects. The second is the large ego-motion in the videos. To deal with shadows, we build up an autonomous driving shadow dataset and design a deep neural network to detect shadows automatically. To deal with large ego-motion, 
we take advantage of the multi-source data, in particular the 3D data, in autonomous driving. More specifically, the geometric relationship between frames is incorporated into an inpainting deep neural network to produce high-quality structurally consistent video output. Experiments show that our method
outperforms other state-of-the-art (SOTA) object removal algorithms,
reducing the RMSE by over $19\%$.

\end{abstract}

\section{Introduction}\label{section:introduction}

\begin{figure}[t]
	\centering  
	\includegraphics[width=\columnwidth]{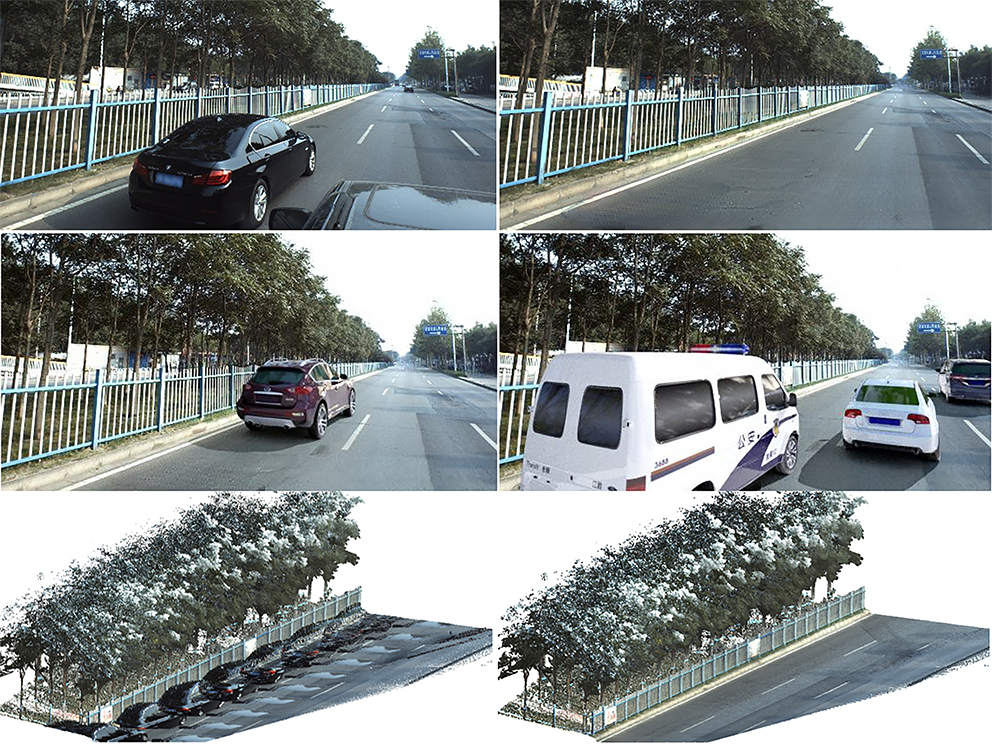}
	\caption{1st row shows the source image and inpainted one from a video. 2nd row shows the usage of inpainting in data augmentation and simulator. With the inpainted background, the vehicle can be moved or inserted to synthesize new traffic images. 3rd row shows inpainted videos are used to yield 3D model with clean texture. } 
	\label{fig:title_image}  
\end{figure}

With the explosive growth of AI robotic techniques, especially the autonomous driving (AD) vehicles, countless images or videos as long as other sensor data are captured daily. 
To fuel the learning-based AI algorithms (such as perception, scene parsing, planning) in those intelligence systems, a large number of annotated data are still in great demand. Thus, building virtual simulators for saving massive efforts on labeling and processing the captured data are essential to make the data best used for various AD applications~\cite{alhaija2018augmented,seif2016autonomous}. One basic procedure in those applications is removing the unwanted or hard-to-annotate parts of the raw data, a.k.a the object removal or image/video inpainting. As shown in Figure~\ref{fig:title_image}, with the developed simulation system in~\cite{li2019aads}, the background image obtained by removing the foreground vehicles can be used to  synthesize  new traffic images with annotations or reconstruct 3D road models with clean textures, which is one of the desirable ways for data augmentation.

The image inpainting problem has been widely investigated, which also forms the basis of video inpainting. 
Technically, image inpainting algorithms either utilize similar patches in the current image to fill the hole by the optimization-based methods or directly hallucinate from training images by the learning-based methods. Recently, the CNNs, especially GANs, hugely advanced the image inpainting technique~\cite{PathakKDDE16,IizukaS017,yu2018free}, yielding visually plausible and impressive results. 
However, directly applying image inpainting techniques to videos suffers from jittering and inconsistency. Thus, different kinds of temporal constraints are introduced in recent video inpainting approaches~\cite{huang2016temporally,xu2019deepflow}, whose core is jointly estimating optical flow and inpainting color.

\begin{figure}[t]
	\centering  
	\includegraphics[width=1\columnwidth]{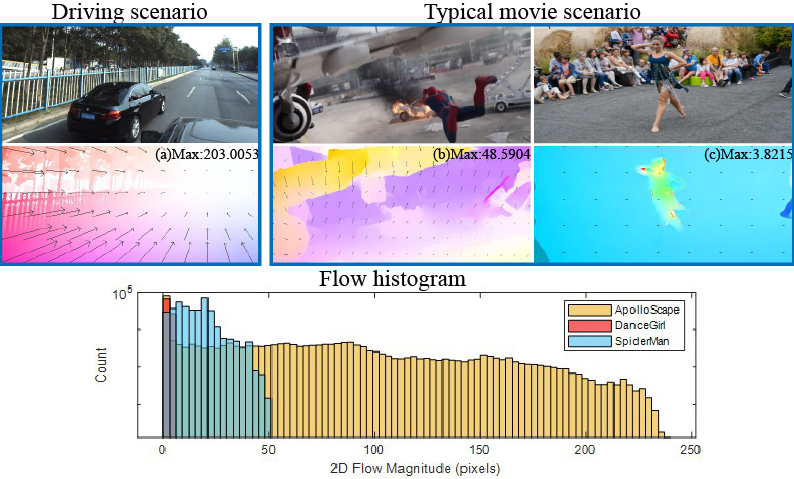} 
	\caption{Visualized flows comparison. The first two rows show RGB images and visualized flow maps on which the arrows represent the flow directions and magnitudes from different videos. (a) is from our dataset while (b)(c) are from other papers~\cite{xu2019deepflow,huang2016temporally}. The last column is the flow histogram. It can be seen that the camera motion of our data is quite large.  } 
	\label{fig:flow_vis_comparison}  
\end{figure}

Even several video inpainting systems have been proposed in the very close recent, their target scenarios are usually with only small camera ego-motion in the behind of foreground objects movements, where the flow between frames are easy to estimate. Unfortunately, the videos captured by AD vehicles have large camera ego-motion (Figure~\ref{fig:flow_vis_comparison} shows the statistics comparison of the optical flows). In addition, the camera ego-motion is usually moving along the heading direction of vehicles, which is also close to camera optical direction. The large vehicle movement and camera projection effect lead to large invisible parts in frames by surrounding vehicles. Moreover, the shadows of foreground objects are either ignored or manually labelled in those system, which does not work in AD scenario obviously.

In this paper, we propose a novel CNN-based object removal algorithm to automatically clean AD videos. The key idea is to utilize 3D convolution to extend the 2D contextual attention in~\cite{yu2018generative} to video inpainting. Specifically, we construct the system using three novel modules: temporal warping, 3D feature extractor and 3D feature assembler. The first module takes the advantage of multi-sensor data to help inpainting. While the last two modules are used to utilize temporal as well as contextual attention (CA) information for inpainting. Technically, naively combining temporal information and CA module is impractical due to large GPU memory footprint. We solve this problem by decomposing and simplifying the CA procedure. 

With regarding to the shadow problem, which is always overlooked in previous inpainting literature, we propose a learning-based module along with an annotated shadow dataset. Our dataset has 5293 frames, which exceeds the SBU~\cite{m_Vicente-etal-ECCV16}, UCF~\cite{m_Vicente-etal-ICCV15} and ISTD dataset~\cite{Wang_2018_CVPR} w.r.t the size. Furthermore, ours advance those datasets in terms of our temporal consistent shadow annotation. Thus, our dataset could be beneficial to more vision tasks compared with typical shadow datasets, e.g. object tracking refinement, illumination estimation. 

In summary, our contributions are as follows:
\begin{itemize} 
	\item We introduce an end-to-end inpainting network which consists of temporal warping, 3D feature extractor and assembler modules to utilize not only temporal but also contextual attention (CA) information for video inpainting, which is experimentally proven to be efficient to the results.
	
	\item We design an indispensable branch to deal with the shadows in AD videos. Our experiments show that it is a must-have module for high-quality results, and flexible to transfer to other SOTA algorithms.
	
	\item We announce a dataset of shadow annotated video AD videos. To the best of our knowledge, our dataset is the first temporal oriented (video) dataset with the largest number of annotated frames.
\end{itemize}

\section{Related Work} \label{section:related_work}

\subsection{Image Inpainting}
Single image inpainting aims to reconstruct the lost parts in images. Patch-based inpainting methods are developed to better reconstruct the contextual structures in images~\cite{barnes2009patchmatch,telea2004image,sun2005image,hays2007scene,huang2014image}. These methods aimed at finding the best-matching patches with structural similarity to fill the missing regions. 

The emergence of deep learning inspires recent works to investigate various deep architectures for image inpainting. 
Learning-based image inpainting directly learns a mapping to predict the missing information~\cite{XieXC12,KohlerSSH14,RenXYS15}. By interpreting images as samples from a high-dimensional probability distribution, image inpainting can be realized by generative adversarial networks~\cite{Goodfellow14,RadfordMC15,mao2016least,mroueh2017fisher,ArjovskyCB17,PathakKDDE16,IizukaS017}. 

Most recently, Yu~\etal~\cite{yu2018generative} presented a contextual attention mechanism in a generative inpainting framework, which improved the inpainting quality. They further extended it to free-form masks inpainting with gated convolution and SN-PatchGAN~\cite{yu2018free}. 

These methods achieve excellent image inpainting results. Extending them directly to videos is, however, challenging due to the lack of temporal constraints modeling.

\begin{figure*}[!ht] 
	\centering  
	\includegraphics[width=\textwidth]{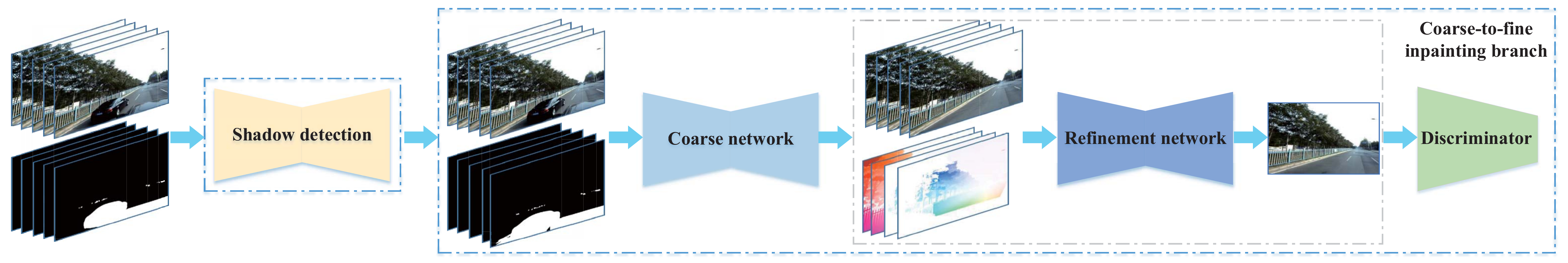} 
	\caption{The pipeline of our approach, which consists of a shadow detection branch and a coarse-to-fine inpainting branch. The shadow detection extends the input object masks to cover their shadows. The inpainting branch inpaints the extended masks in a coarse-to-fine fashion, where the coarse network provides blurry predicts and the refinement network to detailed inpaint the target frame with assembled multi-frames information under the flow guidance. }
	\label{fig:network_structure}
\end{figure*}

\subsection{Video Inpainting}
Video inpainting is generally viewed as an extension of the image inpainting task with larger search space and temporally consistent constraints. 
Extended from patch-based image inpainting, video inpainting algorithms~\cite{wexler2007space,granados2012background,newson_video_2014,ebdelli_video_2015} recovered masked regions by pasting the most similar patches somewhere in the video.
By estimating the optical flow and color jointly, Huang ~\etal~\cite{huang2016temporally} formulated video inpainting problem as a non-parametric patch-based optimization in a temporally coherent fashion. However, the computation time of these methods is still long. In addition, patch-based models still lack modeling distribution of real images, so they fail to recover unseen parts in the video. 

Recently, learning-based video inpainting also gains dramatic improvements. Wang~\etal~\cite{wang2018video} introduced the first learning-based video inpainting framework to predict 3D temporal structure with 3D convolutions implicitly and infer 2D details combined with 3D information. 
Ding~\etal~\cite{ding2019frame} and Chang~\etal~\cite{chang2019free} focused on exploring the spatial and temporal information with convLSTM layers. 
Kim~\etal~\cite{kim_deep_2019} enforced the outputs to be temporally consistent by a recurrent feedback. ~\cite{xu2019deepflow} inpainted all the incomplete optical flows of the video and propagated valid regions to hole regions iteratively.

\section{Approach}\label{section:approach}

Video inpainting aims to remove the foreground objects and synthesize plausible hole-free background. We propose an end-to-end pipeline with a shadow detection branch to remove the objects more thoroughly (Section~\ref{subsection:inpainting}), and a coarse-to-fine inpainting branch (Section~\ref{subsection:inpainting}) to synthesize the background from the information of multi-frames.  
Figure~\ref{fig:network_structure} shows the pipeline of our approach.

\subsection{Shadow Detection Branch}\label{subsection:shadow_detection}

In autonomous videos, the objects are always under complex illuminations, which cast bonded shadows. Simply removing objects with given masks would cause terrible inpainting result. In previous works, shadows are always overlooked. However, in objects removal, dealing with the side effects of objects onto the environment is necessary. Actually, the un-removed shadows not only remain in the result videos leading to moving ghosts, but also heavily misguide the context inpainting in the holes as the shadows are tended to be selected as the best-matching patches.  

One solution to the shadow problem is dilating the mask. However, we experimentally found the increase of hole size would dramatically decrease the inpainting result, since the closer to the original hole, the more content information to guide the inpainting is encoded. In order to automatically generate the inpainting mask with shadow in the as-small-as-possible manner, we propose a shadow detection branch ahead of the practical inpainting blocks.

We construct a classical U-net structure~\cite{ronneberger2015u} for shadow detection. The branch takes RGB images and foreground masks as input and extends the masks with corresponding shadows. The details of this branch can be found in the supplementary material.

\begin{figure*}[!ht] 
	\centering  
	\includegraphics[width=\textwidth]{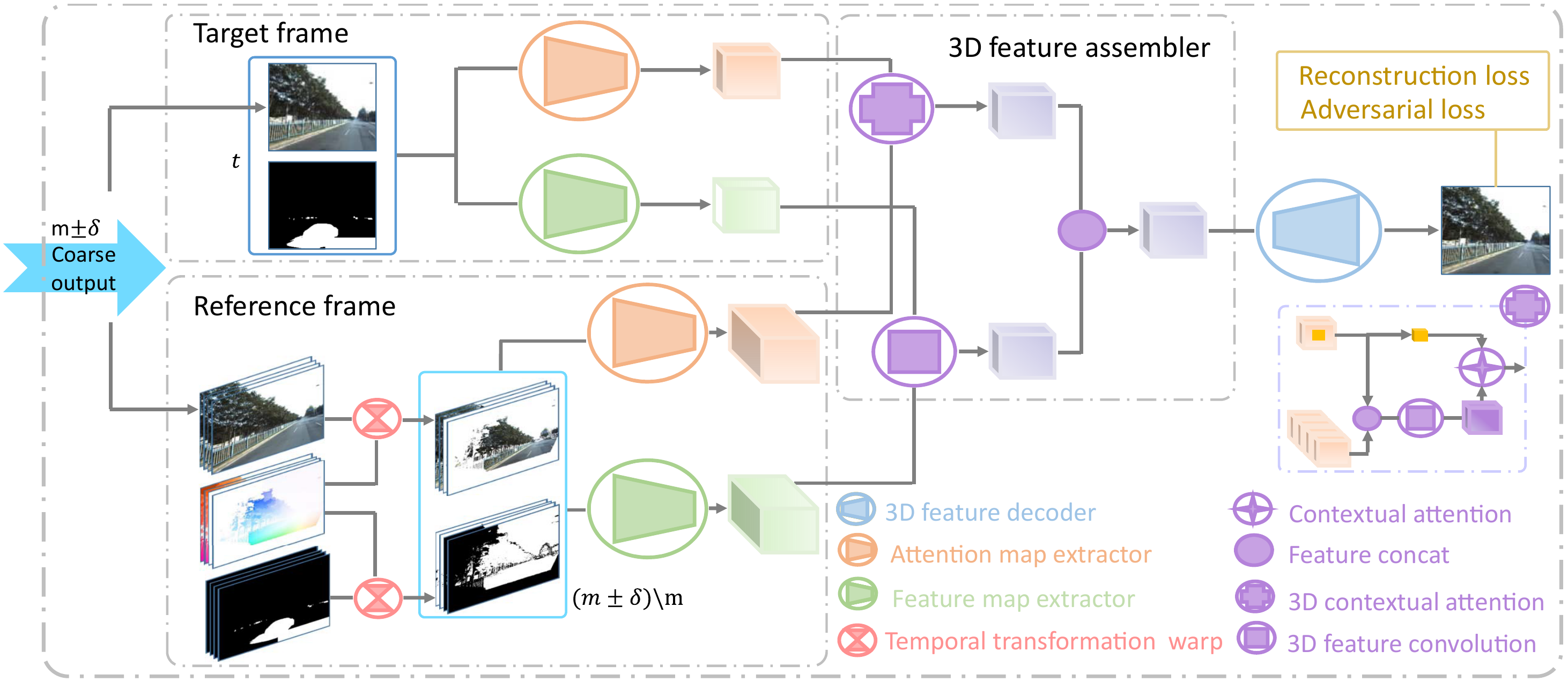} 
	\caption{Architecture of the refinement network. $\delta$ is the interval of frames. $m\pm\delta$ represents all frames from $m-\delta$ to  $m+\delta$. $(m\pm\delta)\setminus m$ represents all frames except $m$. With the guidance of the flows, we align the coarse outputs of each reference frames to the target frame. Then, we extract local contextual attention features and global features using two branches, where all features are later aggregated using a 3D feature assembler to predict target frame. }
	\label{fig:inpainting_branch}
\end{figure*}

\subsection{Coarse-to-fine Inpainting Branch}\label{subsection:inpainting}

Regarding the coarse-to-fine inpainting branch, our method is built on the state-of-the-art single image inpainting network~\cite{yu2018free}. Our method adapts the GAN structure from them, and then introduces the flow constraint from the geometry and temporal consistent constraint to assemble multi-frames together. Concretely, we design three modules to deal with the autonomous driving videos. A temporal warping module is used to aligned different frames to same location, which is experimental important especially when the camera ego-motion is large. Besides, a “3D feature extractor” extracts information of multi-frames and enlarges the searching spaces to supply more alternative matching patches. Then a “3D feature assembler” merges multi-frames to inpaint the target frame, which is effective to improve the temporal consistency and reduce the jittering artifacts among different frames.

Given a video, we utilize every $F$ continuous frames with masks $M$ as input sequences to inpaint the incomplete target frame $I^{m}_{in}$. We use $I_{gt}$ to denote the ground truth of the sequence. The incomplete sequence $I_{in}$ is equal to $I_{gt} * M $, where $*$ is the element-wise multiplying. $U^{m\rightarrow{i} }(i=1,2,\dots,F,i\not = m)$ denotes the flow fields from the target frame to all other ones. $I_{in}$, $M$ and $U$ are input to our coarse-to-fine inpainting branch with the supervision of $I_{gt}$.

As shown in Figure~\ref{fig:network_structure}, the generator $G$ of the inpainting branch is a two-stage coarse-to-fine structure. The coarse network $G_{c}$ provides preliminary and blurry conjectures of the invisible parts, while the refinement network $G_{f}$ refine the results and enhance the details. We follow single image inpainting work~\cite{yu2018generative} to define the structure of the coarse network $G_{c}$. All input frames are processed independently with shared weights of $G_{c}$ in our framework. 

The outputs of coarse inpainting branch $G_{c}(I_{in}, M)$ will be feed into the refinement branch. Figure~\ref{fig:inpainting_branch} shows the detailed structure of the refinement branch, which consists of a temporal warping module, two 3D feature extractors, a 3D feature assembler and a decoder. 

\subsubsection{Temporal warping} 
\begin{figure}[!h]
	\centering  
	\includegraphics[width=\columnwidth]{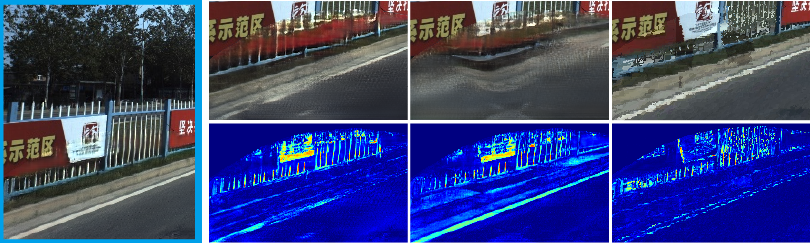}
	\caption{Left to right: our result, result w/o temporal warping, result with predicted flow, result of~\cite{xu2019deepflow} using computed flow but w/o additional blending. The second row shows the visualized difference to our result. }  
	\label{fig:flow_ablation}  
\end{figure}

We propose a \emph{temporal warping module} to transform the coarse network outputs $G_{c}(I_{in}, M)$ to the same camera location using the geometrical guidance. Without this warping module, stacking the frames together directly would lead to the following network blocks to learn the geometry information implicitly. This may work in the typical movie videos in Figure~\ref{fig:flow_vis_comparison}. But for the AD scenarios, the camera ego-motion is relative large. In other words, the receptive field of one convolution kernel on the stacking the frames is small w.r.t the flow caused by camera motion. The second column of Figure~\ref{fig:flow_ablation} shows that the result without temporal warping fails to recover the geometry structure of the scene. Thus, we propose a geometric guided temporal warping module to reduce the effects of large camera ego-motion. Specifically, with $p=(i,j)$ represents the 2D spatial location, all pixels of frames $G_{c}^i $ will be warped to the target frame as follows:
\begin{equation}\label{frames_warping}
\begin{aligned}
G^i_{c}(I_{in},M)'(p) = G^i_{c}(I_{in} , M) & (p + U^{m\rightarrow{i} }) \\
\end{aligned} 
\end{equation}
The warping modlue is implemented with differentiable bi-linear sampling.

\subsubsection{3D feature extractor} There are two different branches to extract feature maps: \emph{contextual attention extractor} and \emph{global feature extractor}. The former one is used to prepare features for the 3D contextual attention block. The latter one is utilized to get an overall impression of the scene and guide the network to hallucinate invisible regions. They take every frame of $G_{c}(I_{in},M)'$ and warped masks $M'$ as input and extract features with gated convolution layers. All frames except the target one are convoluted with shared weights. 

\subsubsection{3D feature assembler} We design a 3D feature assembler to keep the temporal consistency by assembling the outputs of the two extractors. It consists of a multi-frames-to-one 3D contextual attention block and a 3D feature merging block to aggregate all features. 

Although neighboring  frames supply more information to inpaint the holes, there still be invalid regions that can not be seen in any frame. These regions can be inpainted with similar patches in the feature spaces. So we construct the multi-frames-to-one \emph{3D contextual attention block} by combining temporal information and the CA module in~\cite{yu2018free}. The key of this module is to slice the background and foreground features into patches, then choosing the best-matched background patches to inpaint the foreground by similarity scores. To extend to our multi-frames scheme, a straightforward way is to stack background patches in all frames for matching. However, it is impractical since full 3D CA requires large memory footprint. For example, $46080$ background patches will be extracted for our $96*96$ feature map.
Therefore, we simplify the full 3D contextual attention block using a two-stage module. The first stage is a 3D convolution layer of one kernel in depth $F$ to assemble the features of all $F$ frames to aggregated features. The second stage is a CA layer that treat the aggregated features as background and target frame features as foreground, which is used to find matching scores between coarse foreground and background. In the first stage, we can also use bidirectional long short-term memory (LSTM) layer instead of 3D convolution layer. However, it achieves similar result but with more training cost. 

The \emph{3D feature merging block} is also a 3D convolution layer used for the global guidance feature map extractor. The output of 3D feature merging block and 3D contextual attention block will be concatenated together to maintain the global structure information and local patches for inpainting.  

The assembled features are input to the decoder and the incomplete target frame $I^m_{gt}$ are inpainted as $G^m(I_{in},M,U)$. At last, a spectral-normalized Markovian discriminator (SN-PatchGAN) $D$ as~\cite{yu2018free} is used to hallucinate the missing regions in all frames. The details can be referenced in the supplementary material.

\subsection{Loss Function}\label{subsection:loss_function}
The objective function of the video inpainting branch consists of a reconstruction loss and an adversarial loss. The reconstruction loss $L_{g}$ is an $L1$ loss combined with the coarse network and the refinement network. $L_{g}$ is defined as:
\begin{equation}
L_{g}= \alpha ||G_{c}(I_{in}, M)-I_{gt}|| + ||G^m(I_{in},M,U) - I^m_{gt}||
\end{equation}
where $\alpha$ is the balancing parameter and $||\cdot||$ is the $l_{1}$ norm. During the training, gradients only back-propagate at non-object regions. 
The discriminator takes $G^m(I_{in},M,U)$, $I^{m}_{gt}$ as input and outputs a feature map with each value represents the corresponding region in the image is fake or not. The adversarial loss is a hinge loss:
\begin{equation}
\begin{aligned}
L_{D}= & \mathbf{E_{x\sim \mathbf{P}_{data}(x)}}[max(0,1-D(x))] +\\
& \mathbf{E_{z\sim \mathbf{P}_{z}(z)}}[max(0,1+D(z))] \\
L_{G}= & -\mathbf{E_{z\sim \mathbf{P}_{z}(z)}}[D(z)] 
\end{aligned}
\end{equation}
where $x=I^m_{gt}$ and $z$ is the generator output $G^m(I_{in},M,U)$.

For the shadow detection branch, a weighted binary cross entropy loss same as~\cite{xie2015holistically} is adopted. Please refer to the supplementary material for details.

\subsection{Data Generation}\label{subsection:data_generation}

Based on the AD dataset, we prepare two kinds of data in order to feed our pipeline: 
\begin{itemize}
	\item shadow dataset: a number of images with object shadows are manually annotated for training shadow detection branch.
	\item inpainting data: the inpainting data are augmented in two ways: 1) synthesized images using AD simulator with realistic objects and shadows, which is used for evaluation and training (both shadow detection branch and inpainting branch), 2) large number of images with generated temporal consistent masks for training inpainting branch.
\end{itemize}

\begin{figure}[h]
	\begin{subfigure}{\columnwidth}
		\includegraphics[width=\columnwidth]{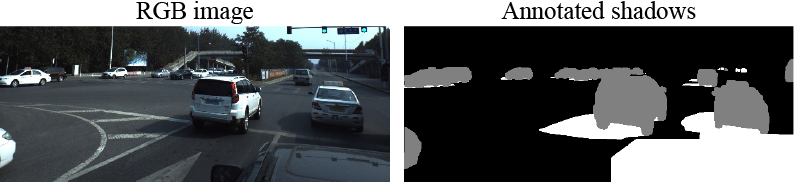}
		\caption{An example image with annotated shadows.}
	\end{subfigure}
	\begin{subfigure}{\columnwidth}
		\includegraphics[width=\columnwidth]{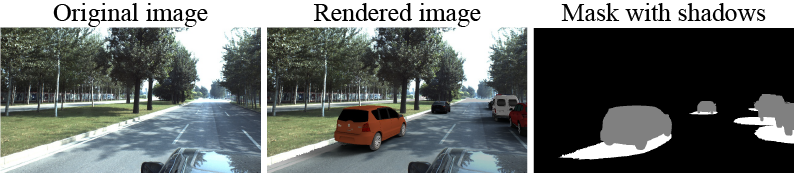}
		\caption{An example synthetic image using AD simulator.}
	\end{subfigure}
	\begin{subfigure}{\columnwidth}
		\includegraphics[width=\columnwidth]{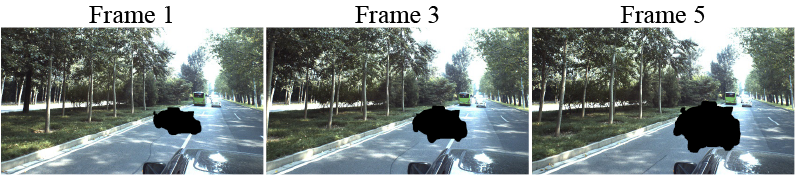}
		\caption{A training image sequence with temporal consistent masks.}
	\end{subfigure}
	\caption{Data generation}
	\label{fig:data_generation}
\end{figure}

\subsubsection{Shadow Dataset}
To train our shadow-detection branch, we annotated a shadow detection dataset including $5293$ images. Shadow regions of the foreground objects especially the cars are labelled manually. Figure~\ref{fig:data_generation}(a) shows an example of the dataset. As the shadow areas provide implicit hints for lighting sources, illumination conditions and scene geometry, shadow detection is helpful for scene understanding and geometry perception. Cucchiara~\etal~\cite{Cucchiara2001Improving}and Mikic~\etal~\cite{Mikic2000Moving} have shown that some vision tasks, like efficient object detection and tracking (which is the key topic in the autonomous driving), can be beneficial from shadow removal.To best of our known, our annotated shadow dataset is the first video objects shadow dataset and supplies comparable data with the largest shadow dataset. There are three common shadow detections: the largest SBU Shadow Dataset with 4727 images, the UCF Shadow Dataset with 221 and the ISTD dataset with 1870 images while our dataset provides 5293 images. In these datasets, all shadows are marked without caring about foreground objects or background and the images in the dataset are all independent. Different from them, our dataset only annotates the shadows of the foreground objects and supplies the temporal consistent annotations of videos, which can be beneficial to some vision tasks, like efficient object detection and tracking. This dataset will be released to the public with the paper.

\begin{figure*}[!ht]
	\centering  
	\includegraphics[width=\textwidth]{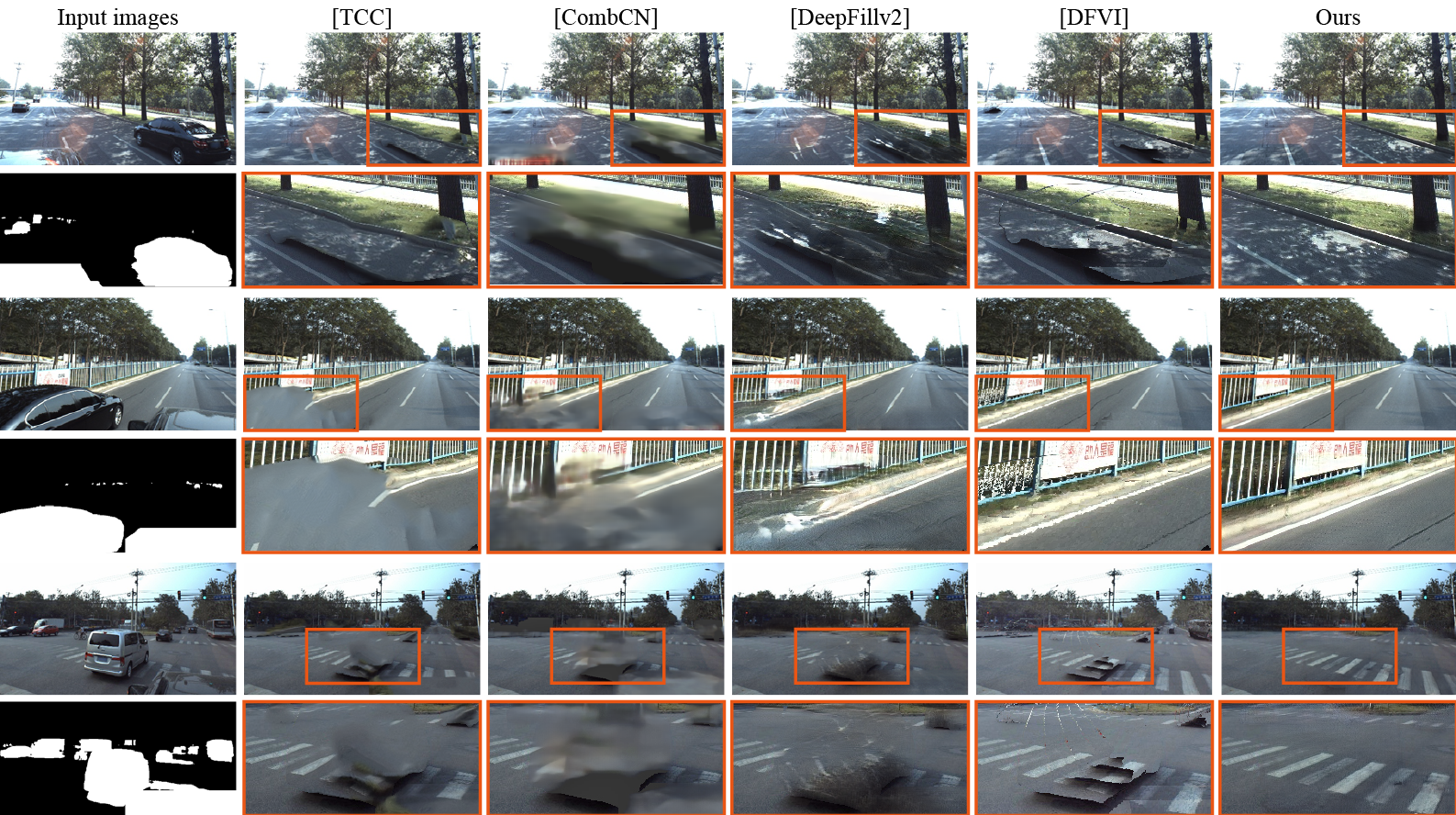}
	\caption{Comparisons with existing methods. The regions of the orange boxes are enlarged for details. }  
	\label{fig:comparison_with_existing_methods_true}  
\end{figure*}

\subsubsection{Inpainting Data}
Synthetic images are generated using the AD simulator AADS~\cite{li2019aads}. AADS is a close-loop AD simulator which can simulate traffic flow, LiDAR and color images. The reason of choosing AADS is that AADS could augment AD scenes with realistic objects under estimated illuminations. Thus, our synthetic images can be used as the supplementary of annotated shadow dataset. Furthermore, synthetic images are also used for quantitative evaluation, since the ground truth of video inpainting is not easy to obtain. Actually, there is no method who used the clean backgrounds after removing the real objects as the ground truth. Existing methods were only evaluated visually or by user study. Specifically, we run the simulator on ApolloScape images with few vehicles. Benefiting from the simulator's lighting estimation module and traffic simulation backend, the synthetic images are with environment consistent shadows and temporal consistent object masks. Figure~\ref{fig:data_generation}(b) shows an example of our synthetic data.

As extra bundled information, such as HD maps with lanes, is required to run the AD simulators. Synthetic images using simulator are very limited. Thus, for those images without extra information, we generate temporal consistent holes by warping the masks onto different frames and add random displacements to simulate object moving. Such temporal consistent masks are used as the main source of training data, which is shown in Figure~\ref{fig:data_generation}(c).

\section{Experiments and Results} \label{section:results}

\subsection{Implementation Details}
We focus on inpainting the street-view videos, especially those coming from AD datasets. In our paper, we use the ApolloScape~\cite{huang2018apolloscape} for experiments. The ApolloScape is a large-scale AD dataset which contains rich labeling including per-pixel semantic labelling, instance segmentation, projected depth images and corresponding camera poses. Please note that ApolloScape provides depth maps of static background, so the flows $U$ can be directly computed from depth without flow inpainting algorithm like~\cite{xu2019deepflow}. 
In our implementation, number of frames in one sample sequence is set to $5$. The original images in ApolloScape are downsampled to $1/4$. Total 16373 samples of training sequence are generated. During the training, the images are bottom cropped to $562*226$ and then randomly cropped to $384*192$ around the generated holes. The network is trained with an Adam optimizer for 210k iterations, whose learning rate is $0.0001$ and batch size is 8. All the experiments are implemented with Tensorflow \& PaddlePaddle and performed on 4 NVIDIA Tesla P40. During the testing, images are center cropped into $560*448$.

\subsection{Comparisons with existing methods}
\subsubsection{Baselines}
We compare our approach with four different state-of-the-art methods, from single image inpainting to classical video inpainting and video inpainting with deep neural networks. The baselines are as following:
\begin{itemize}
	\item Deepfillv2~\cite{yu2018free} is the state-of-art single image inpainting algorithm. We re-implement the method by extending the released code of~\cite{yu2018generative}.
	\item TCC~\cite{huang2016temporally} is a classical optimization-based video inpainting algorithm, which released MATLAB code for comparison. 
	\item CombCN~\cite{wang2018video} is the first work to utilize deep neural networks for video inpainting. We re-implement the architecture with Tensorflow, and modified the inputs to be consistent with our data.
	\item DFVI~\cite{xu2019deepflow} is the most recent video inpainting algorithm. It learns to inpaint flow maps as well as holes. We use the release code for comparison.
\end{itemize}
All neutral network are re-trained with our generated data. For TCC and DFVI, we use our generated flow as their predicted flow for a fair comparison. 

\begin{figure}[ht]
	\centering  
	\includegraphics[width=\columnwidth]{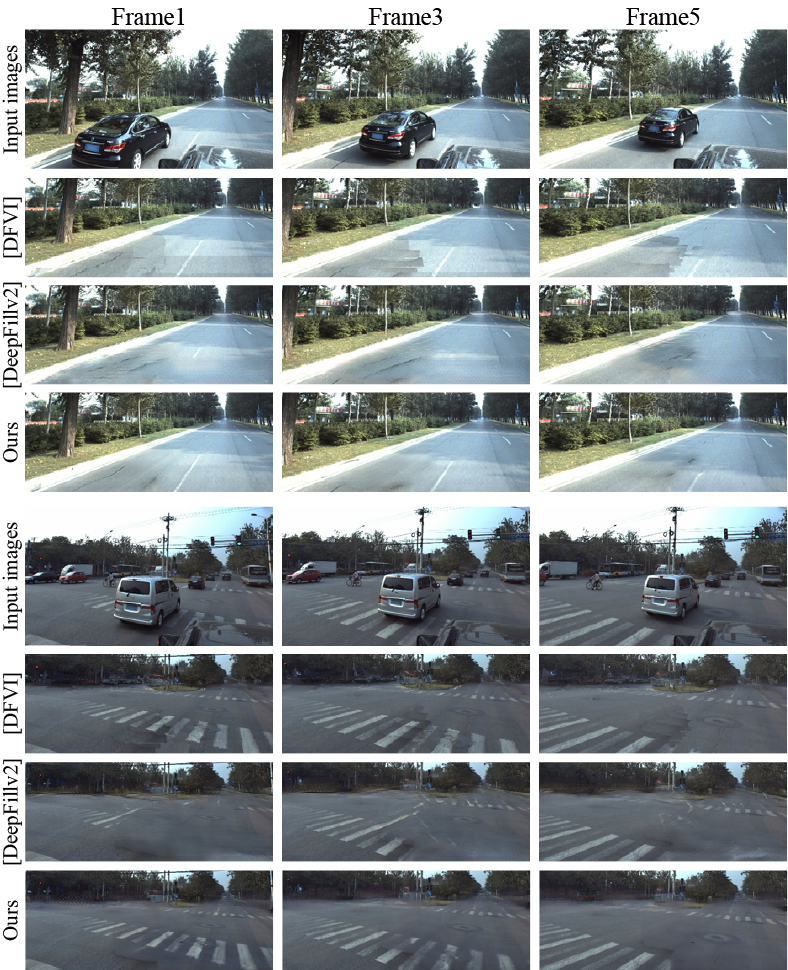}
	\caption{Comparisons with existing methods. To remove the impacts of the shadows and emphasize the temporal consistency, shadow detection is introduced to all methods. }  
	\label{fig:comparison_of_sequence} 
\end{figure}

\subsubsection{Comparison and Analysis}
Figure~\ref{fig:comparison_with_existing_methods_true} shows visual comparisons on the single frame of inpainted videos between the baseline methods and our method. One significant improvement of our method is the shadows and ghosts eliminating. With shadow-aware branch, our method can remove the moving objects completely and obtain cleaner backgrounds.

Comparing to the baseline methods, our method achieve better results with less artifacts. Deepfillv2 may fail when the holes are large, as it is hard to keep geometric structure for this single image inpainting framework. TCC method follows joint flow and color optimization. Even using our computed flow as reliable initialization, TCC still produces unacceptable results with mismatched boundaries shown in the second column of Figure~\ref{fig:comparison_with_existing_methods_true}. CombCN relies on typical 3D convolution to implicitly maintain the temporal information which is not enough under large camera ego-motion. Thus the results of CombCN is short of structure information. As the newest SOTA video inpainting method based on neural networks, DFVI performs well on maintain the geometry structure of the scenes. However, the brightness and illumination of frames always change a lot even in one sequence when the camera motion is large. Thus, propagating patches directly, which is used by DFVI, cannot yield smooth blended results. Especially when the flow is not very accurate, the results are noisy. This usually appears on thin objects like the fences in the second rows of Figure~\ref{fig:comparison_with_existing_methods_true}. Besides, the propagation used by DFVI is very time consuming especially when the hole regions can not be borrowed from other frames directly. It will inpaint the key-frames and propagates them to all frames iteratively. In terms of one sequence of 175 frames in our evalation data, the runtime of DFVI is 20 minutes while our method can inpaint them in 40 seconds.

In our method, with the temporal warping module, the geometry structure can be well maintained even when camera ego-motion is large. With the 3D feature assembler, the features of different frames can be blended smoothly. With the shadow-aware branch, the moving-shadow artifacts can be solved. Thus, our method yield the best visual results comparing to existing methods in AD videos. More results can be found in the supplementary materials.

To quantitative compare our method with other methods, we utilize five metrics for the evaluations: mean absolute error(MAE), root mean squared error(RMSE), peak signal to noise ratio(PSNR), structural similarity index(SSIM) and temporal warping root mean squared error(TWE). We calculate the TWE by warping one inpainted frame to next frame and computing the RMSE on the valid regions. The TWE is applied on different frames to evaluate the temporal consistency. Note that those metrics are evaluated only on the inpainted hole regions. Table ~\ref{tab:comparisons_with_existing_methods} shows the evaluation results of the baseline methods and our method. Note that our method outperforms others on all the metrics.

\begin{table}
	\begin{center}
		\small
		\begin{tabular}{c|c c|c c|c}
			\hline
			Method & MAE & RMSE & PSNR & SSIM & TME \\
			\hline
			TCC & 22.595 & 31.322 & 32.332 & 0.9657 & 29.490 \\
			CombCN & 19.952 & 28.059 & 33.332 & 0.9686 &25.848 \\
			Deepfillv2 & 18.725 &28.004 & 33.053 & 0.9626 & 30.260\\
			DFVI & 23.005 & 35.110 & 31.282 & 0.9674 & 26.514 \\
			Ours & \textbf{15.143} & \textbf{22.611} & \textbf{34.435} & \textbf{0.9697} & \textbf{24.822}\\
			\hline
		\end{tabular}
	\end{center}
	\caption{Comparison with different methods. Our method outperforms others on all metrics.}
	\label{tab:comparisons_with_existing_methods}
\end{table}

To remove the effects of shadows and compare the video inpainting branch only, we also add the shadow detection branch to the baseline methods for comparison. Figure~\ref{fig:comparison_of_sequence} shows some frames of the inpainted videos. As this figure shows, our method yields better results with temporal consistency, which benefits from the guidance of the geometric information from multi-frames features. Please refer to the supplementary video for a better view and more results.

\begin{table}
	\begin{center}
		\small
		\begin{tabular}{c|c c}
			\hline
			Method & MAE & RMSE \\
			\hline 		
			Deepfillv2 & 19.799 & 29.544\\
			Deepfillv2 + shadow detection & 18.725 & 28.004 \\
			Ours - w/o temporal warping & 16.202 & 24.437  \\
			Ours - w/o contextual attention & 16.236 & 24.051 \\
			Ours - w/o shadow detection & 15.414 & 23.217  \\
			Ours & \textbf{15.143} & \textbf{22.611}  \\
			\hline
		\end{tabular}
	\end{center}
	\caption{Evaluation metrics of ablation study. }
	\label{tab:ablation_study}
\end{table}

\subsection{Ablation Study}
To explore the effects of different parts of our algorithm, we conduct several ablation experiments about the multi-frames-to-one scheme, the temporal warping module and the contextual attention in 3D feature assembler and shadow detection branch. Table~\ref{tab:ablation_study} shows the evaluation metrics of the ablation study. Adding the multi-frames-to-one scheme without temporal warping module reduces the MAE from $18.725$ to $16.202$ as the network could find matching patches from more frames. Besides, adding the temporal warping module can also gain further promotion of the metrics. It can utilize the geometric information to guide the inpainting process explicitly. Features of similar objects are aligned together, making it easier to find the matching patches. Please refer to Figure~\ref{fig:flow_ablation} for the image results. Removing the contextual attention in 3D feature assembler leads to larger errors. The ablation studies of the temporal warping module and the contextual attention in Table~\ref{tab:ablation_study} show that objects removal can be beneficial from the combination of temporal information and the contextual attention. 

\begin{figure}
	\centering  
	\includegraphics[width=\columnwidth]{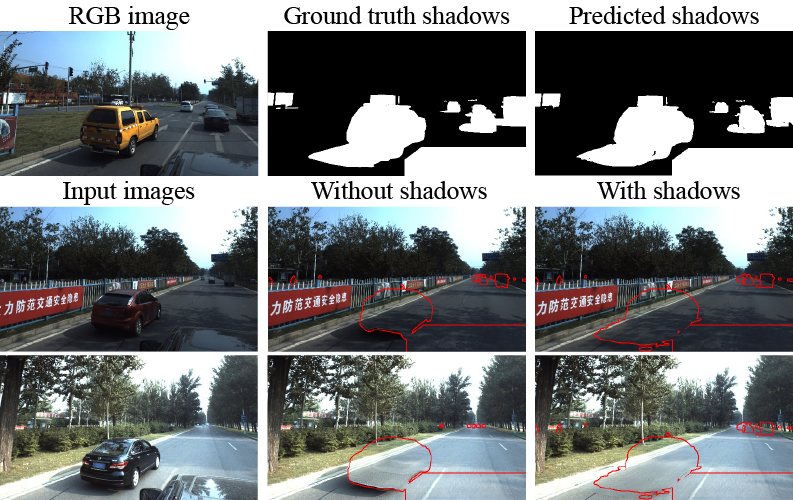}
	\caption{The first row shows the results of shadow detection network. The second and third rows show inpainting results with/without shadow detection. The removal regions are marked with red boundaries.}
	\label{fig:inpainting_results_w-wo_shadow_detection}  
\end{figure}

The shadow detection branch is one of the most effective parts to get the clean inpainted background. The first row of Figure~\ref{fig:inpainting_results_w-wo_shadow_detection} shows some examples of our shadow detection network's predicts. Removing shadows along with cars has a shortcoming that the holes are enlarged. It is acknowledged that larger holes are harder to inpaint. However, it also can reduce the difficulty to find the matching patches and remove the moving shadow ghosts in videos. With the predicted shadow maps, the influence of wrong matching patches with different brightness can be reduced and the ghosts of the shadows will be removed. From the Table~\ref{tab:ablation_study}, adding shadow detection branch improves the performance. 

Besides the reducing of the inpainting errors, the most obvious improvement is the visual effects. As the bottom two rows of Figure~\ref{fig:inpainting_results_w-wo_shadow_detection} shows, with the shadow detection branch, the foreground objects could be removed more thoroughly. Shadow detection branch is a unified block that could be added to any AD video inpainting algorithm as a pre-processing operation.

\subsection{The Generalization Ability}
\begin{figure}
	\centering  
	\includegraphics[width=\columnwidth]{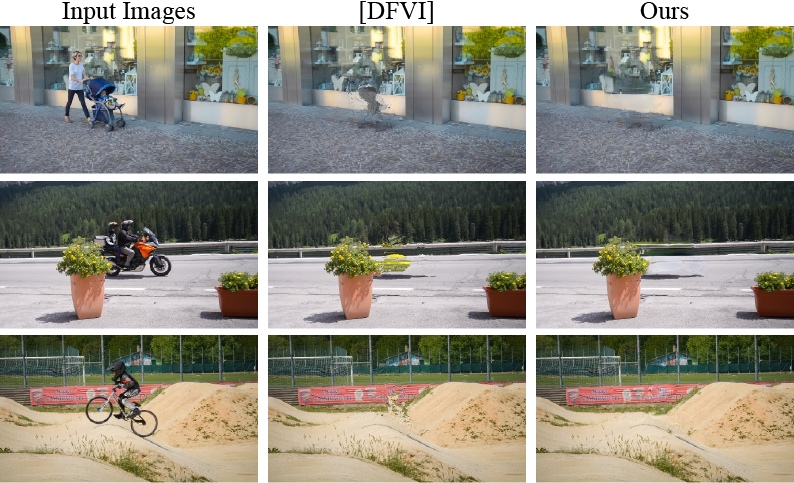}
	\caption{Comparison with DFVI on the DAVIS dataset.}
	\label{fig:davis_results}  
\end{figure}

To evaluate the generalization ability of our method, we train our model on the DAVIS dataset~\cite{Perazzi2016davis}. The results are shown in Figure~\ref{fig:davis_results}. The inaccurate predicted flows misguide the inpainting in DFVI while our results can preserve the geometry structures of the backgrounds.

\section{Conclusion}
In this paper, we present a shadow-aware video inpainting pipeline for street-view foreground objects removal in AD with large camera motion. We use a multi-frames-to-one scheme with the geometry guidance to aggregate information of multi-frames and keep the temporal consistency. A unified shadow detection branch is adopted for removing the shadow ghosts and reducing the impacts of redundant patches. The first foreground objects shadow detection dataset focusing on AD will be open source. In the experiments, we propose a new evaluation method for objects removal when there is no ground truths. The experiments demonstrates that our methods could reduce the artifacts and reconstruct clean background images. In the future, we will investigate the method to reduce the running time for real-time application and improve the performances of regions can not be borrowed from other frames.
\section{Acknowledgements}
Weiwei Xu is partially supported by NSFC (No. 61732016) and the fundamental research fund for the central universities. Jinhui Yu is partially supported by NSFC (No. 61772463). We also thank the reviewers and all the people
who offered help.
{
\small
\bibliography{autoremover}
\bibliographystyle{aaai}
}

\end{document}